\def\year{2017}\relax
\documentclass[letterpaper]{article}

\def \tr{{\mathrm{tr}}}
\def \bbR{{\mathbb{R}}}
\def \cH{{\mathcal{H}}}
\def \cD{{\mathcal{D}}}
\def \sS{{\mathscr{S}}}
\def \sT{{\mathscr{T}}}

\def \bA{{\mathbf{A}}}
\def \bD{{\mathbf{D}}}
\def \bH{{\mathbf{H}}}
\def \bI{{\mathbf{I}}}
\def \bK{{\mathbf{K}}}
\def \bM{{\mathbf{M}}}
\def \bX{{\mathbf{X}}}
\def \bZ{{\mathbf{Z}}}
\def \bL{{\mathbf{L}}}

\def \bW{{\mathbf{W}}}

\def \bk{{\mathbf{k}}}

\def \bw{{\mathbf{w}}}
\def \bx{{\mathbf{x}}}
\def \by{{\mathbf{y}}}
\def \bz{{\mathbf{z}}}

\usepackage{float}
\usepackage{algorithm}
\usepackage[noend]{algpseudocode}
\usepackage{colortbl}
\usepackage{tablefootnote}
\usepackage{amsfonts}
\usepackage{mathrsfs}
\usepackage{amsmath}
\usepackage{mathtools}
\usepackage{tabularx}
\usepackage{subfig}
\usepackage{array, multirow}
\usepackage{hhline}
\usepackage{multicol}

\usepackage{graphicx}
\usepackage{graphics}
\usepackage{amsmath,amssymb} 
\usepackage{color}
\usepackage{epstopdf}

\usepackage{xcolor,colortbl}
\definecolor{Gray}{gray}{0.85}
\newcolumntype{g}{>{\columncolor{Gray}}c}

\usepackage{aaai17}
\usepackage{times}
\usepackage{helvet}
\usepackage{courier}
\begin{document}
%
\title{Model Selection with Nonlinear Embedding for Unsupervised Domain Adaptation}
\author{Hemanth Venkateswara, Shayok Chakraborty, Troy McDaniel, Sethuraman Panchanathan\\
Center for Cognitive Ubiquitous Computing, Arizona State University, Tempe, AZ, USA\\
\{hemanthv, shayok.chakraborty, troy.mcdaniel, panch\}@asu.edu\\
}
\maketitle
\begin{abstract}
Domain adaptation deals with adapting classifiers trained on data from a source distribution, to work effectively on data from a target distribution. 
In this paper, we introduce the Nonlinear Embedding Transform (NET) for unsupervised domain adaptation. 
The NET reduces cross-domain disparity through nonlinear domain alignment. 
It also embeds the domain-aligned data such that similar data points are clustered together. 
This results in enhanced classification. 
To determine the parameters in the NET model (and in other unsupervised domain adaptation models), we introduce a validation procedure by sampling source data points that are similar in distribution to the target data. 
We test the NET and the validation procedure using popular image datasets and compare the classification results across competitive procedures for unsupervised domain adaptation. 
\end{abstract}

\section{Introduction}
There are large volumes of unlabeled data available online, owing to the exponential increase in the number of images and videos uploaded online. 
It would be easy to obtain labeled data if trained classifiers could predict the labels for unlabeled data. 
However, classifier models do not perform well when applied to unlabeled data from different distributions, owing to domain-shift \cite{torralba2011unbiased}. 
Domain adaptation deals with adapting classifiers trained on data from a source distribution, to work effectively on data from a target distribution \cite{pan2010survey}. 
Some domain adaptation techniques assume the presence of a few labels for the target data, to assist in training a domain adaptive classifier \cite{aytar2011tabula,duan2012domain,hoffman2013}. 
However, real world applications need not support labeled data in the target domain and adaptation here is termed as unsupervised domain adaptation. 

Many of the unsupervised domain adaptation techniques can be organized into \textit{linear} and \textit{nonlinear} procedures, based on how the data is handled by the domain adaptation model. 
A \textit{linear} domain adaptation model performs linear transformations on the data to align the source and target domains or, it trains an adaptive linear classifier for both the domains; for example a linear SVM \cite{bruzzone2010domain}. 
\textit{Nonlinear} techniques are deployed in situations where the source and target domains cannot be aligned using linear transformations.  
These techniques apply nonlinear transformations on the source and target data in order to align them.
For example, Maximum Mean Discrepancy (MMD) is applied to learn nonlinear representations, where the difference between the source and target distributions is minimized \cite{pan2011domain}. 
Even though nonlinear transformations may align the domains, the resulting data may not be conducive to classification. 
If, after domain alignment, the data were to be clustered based on similarity, it can lead to effective classification. 
\begin{figure}[t]
\centering
		\includegraphics[width=0.80\linewidth]{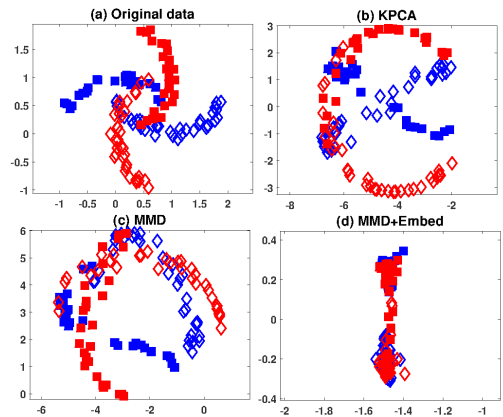}
\caption{\scriptsize{(Best viewed in color) Two-moon binary classification problem with ${\color{blue}\text{source data in blue}}$ and ${\color{red}\text{target data in red}}$. We assume the target labels are unknown. (a) Original data, (b) KPCA aligns the data along nonlinear directions of maximum variance, (c) MMD aligns the two domains, (d) MMD+Similarity-based Embedding aligns the domains and clusters the data to ensure easy classification.}}
\label{Fig:Toy}
\end{figure}
We demonstrate this intuition through a binary classification problem using a toy dataset. 
Figure (\ref{Fig:Toy}a), displays the source and target domains of a two-moon dataset. 
Figure (\ref{Fig:Toy}b), depicts the transformed data after KPCA (nonlinear projection). 
In trying to project the data onto a common `subspace', the source data gets dispersed. 
Figure (\ref{Fig:Toy}c), presents the data after domain alignment using Maximum Mean Discrepancy (MMD). 
Although the domains are now aligned, it does not necessarily ensure enhanced classification. 
Figure (\ref{Fig:Toy}d), shows the data after MMD and similarity-based embedding, where data is clustered based on class label similarity. 
Cross-domain alignment along with similarity-based embedding, makes the data classification friendly.

In this work, we the present the Nonlinear Embedding Transform (NET) procedure for unsupervised domain adaptation. 
The NET performs a nonlinear transformation to align the source and target domains and also cluster the data based on label-similarity. 
The NET algorithm is a spectral (eigen) technique that requires certain parameters (like number of eigen bases, etc.) to be pre-determined. 
These parameters are often given random values which need not be optimal \cite{pan2011domain,long2013transfer,long2014transfer}. 
In this work, we also outline a validation procedure to fine-tune model parameters with a validation set created from the source data. 
In the following, we outline the two main contributions in our work: 
\begin{itemize}
\item Nonlinear embedding transform (NET) algorithm for unsupervised domain adaptation.
\item Validation procedure to estimate optimal parameters for an unsupervised domain adaptation algorithm.  
\end{itemize}
We evaluate the validation procedure and the NET algorithm using 7 popular domain adaptation image datasets, including object, face, facial expression and digit recognition datasets. 
We conduct 50 different domain adaptation experiments to compare the proposed techniques with existing competitive procedures for unsupervised domain adaptation.

\section{Related Work}
For the purpose of this paper, we discuss the relevant literature under the categories \textit{linear} domain adaptation methods and \textit{nonlinear} domain adaptation methods.
A detailed survey on transfer learning procedures can be found in \cite{pan2010survey}. A survey of domain adaptation techniques for vision data is provided by \cite{patel2015visual}. 

The Domain Adaptive SVM (DASVM) \cite{bruzzone2010domain}, is an unsupervised method that iteratively adapts a linear SVM from the source to the target. 
In recent years, the popular unsupervised linear domain adaptation procedures are Subspace Alignment (SA) \cite{fernando2013unsupervised}, and the Correlation Alignment (CA) \cite{sun2015return}. 
The SA algorithm determines a linear transformation to project the source and target to a common subspace, where the domain disparity is minimized. 
The CA is an interesting technique which argues that aligning the correlation matrices of the source and target data is sufficient to reduce domain disparity. 
Both the SA and CA are linear procedures, whereas the NET is a nonlinear method. 

Although deep learning procedures are inherently highly nonlinear, we limit the scope of our work to nonlinear transformation of data that usually involves a positive semi-definite kernel function. 
Such procedures are closely related to the NET. 
However, in our experiments, we do study the NET with deep features also. 
The Geodesic Flow Kernel (GFK) \cite{gong2012geodesic}, is a popular domain adaptation method, where the subspace spanning the source data is gradually transformed into the target subspace along a path on the Grassmann manifold of subspaces. 
Spectral procedures like the Transfer  Component Analysis (TCA) \cite{pan2011domain}, the Joint Distribution Alignment (JDA) \cite{long2013transfer} and Transfer Joint Matching (TJM) \cite{long2014transfer}, are the most closely related techniques to the NET. 
All of these procedures involve a solution to a generalized eigen-value problem in order to determine a projection matrix to nonlinearly align the source and target data.
In these spectral methods, domain alignment is implemented using variants of MMD, which was first introduced in the TCA procedure. 
JDA introduces joint distribution alignment which is an improvement over TCA that only incorporates marginal distribution alignment. 
The TJM performs domain alignment along with instance selection by sampling only relevant source data points. 
In addition to domain alignment with MMD, the NET algorithm implements similarity-based embedding for enhanced classification. 
We also introduce a validation procedure to estimate the model parameters for unsupervised domain adaptation approaches. 

\section{Domain Adaptation With Nonlinear Embedding}
In this section, we first outline the NET algorithm for unsupervised domain adaptation.
We then describe a cross-validation procedure that is used to estimate the model parameters for the NET algorithm. 

We begin with the problem definition where we consider two domains; source domain $\sS$ and target domain $\sT$. 
Let $\cD_s = \{(\bx_i^s, y_i^s)\}_{i=1}^{n_s} \subset \sS$ be a subset of the source domain and $\cD_t = \{(\bx_i^t, y_i^t)\}_{i=1}^{n_t} \subset \sT$ be the subset of the target domain. 
Let \begin{footnotesize}$\bX_S = [\bx_1^s, \ldots, \bx_{n_s}^s] \in \bbR^{d\times n_s}$\end{footnotesize} and \begin{footnotesize}$\bX_T = [\bx_1^t, \ldots, \bx_{n_t}^t] \in \bbR^{d\times n_t}$\end{footnotesize} be the source and target data points respectively. 
Let $Y_S = [y_1^s, \ldots, y_{n_s}^s]$ and $Y_T = [y_1^t, \ldots, y_{n_t}^t]$ be the source and target labels respectively. 
Here, $\bx_i^s$ and $\bx_i^t$ $\in \bbR^d$ are data points and $y_i^s$ and $y_i^t$ $\in \{1,\ldots,C\}$ are the associated labels. 
We define \begin{footnotesize}$\bX \coloneqq [\bx_1, \ldots, \bx_n] = [\bX_S, \bX_T]$\end{footnotesize}, where $n = n_s + n_t$. 
The problem of domain adaptation deals with the situation where the joint distributions for the source and target domains are different, i.e. \begin{footnotesize}$P_S(X,Y) \neq P_T(X,Y)$\end{footnotesize}, where $X$ and $Y$ denote random variables for data points and labels respectively. 
In the case of unsupervised domain adaptation, the labels $Y_T$ are unknown. 
The goal of unsupervised domain adaptation is to estimate the labels of the target data $\hat{Y}_T = [\hat{y}_1^t, \ldots, \hat{y}_{n_t}^t]$ corresponding to $\bX_T$ using $\cD_s$ and $\bX_T$. 

\subsection{Nonlinear Domain Alignment}
A common procedure to align two datasets is to first project them to a common subspace. 
Kernel-PCA (KPCA) estimates a nonlinear basis for such a projection. 
In this case, data is internally mapped to a high-dimensional (possibly infinite-dimensional) space defined by \begin{footnotesize}$\Phi(\bX) = [\phi(\bx_1), \ldots, \phi(\bx_n)]$\end{footnotesize}. 
\begin{footnotesize}$\phi:\bbR^d \rightarrow \cH$\end{footnotesize} is the mapping function and $\cH$ is a RKHS (Reproducing Kernel Hilbert Space). 
The dot product between the mapped vectors $\phi(\bx)$ and $\phi(\by)$, is estimated by a positive semi-definite (psd) kernel, $k(\bx,\by) = \phi(\bx)^\top\phi(\by)$. 
The dot product captures the similarity between $\bx$ and $\by$. 
The kernel similarity (gram) matrix consisting of similarities between all the data points in \begin{footnotesize}$\bX$\end{footnotesize}, is given by, \begin{footnotesize}$\bK = \Phi(\bX)^\top\Phi(\bX) \in \bbR^{n\times n}$\end{footnotesize}. 
The matrix \begin{footnotesize}$\bK$\end{footnotesize} is used to determine the projection matrix \begin{footnotesize}$\bA$\end{footnotesize}, by solving,
\begin{footnotesize}
	\begin{flalign}
		\max_{\bA^\top\bA = \bI} \tr(\bA^\top\bK\bH\bK^\top\bA).
		\label{Eq:kpca}
	\end{flalign}
\end{footnotesize}
Here, $\bH$ is the $n \times n$ centering matrix given by \begin{footnotesize}$\bH = \bI - \frac{1}{n}\mathbf{1}$\end{footnotesize}, where $\bI$ is an identity matrix and $\mathbf{1}$ is a $n \times n$ matrix of 1s. 
$\bA \in \bbR^{n\times k}$, is the matrix of coefficients and the nonlinear projected data is given by \begin{footnotesize}$\bZ = [\bz_1, \ldots, \bz_n] = \bA^\top\bK \in \bbR^{k\times n}$\end{footnotesize}. 
\noindent Along with projecting the source and target data to a common subspace, the domain-disparity between the two datasets must also be reduced. 
We employ the Maximum Mean Discrepancy (MMD) \cite{gretton2009covariate}, which is a standard nonparametric measure to estimate domain disparity. 
We adopt the Joint Distribution Adaptation (JDA) \cite{long2013transfer}, algorithm which seeks to align both the the marginal and conditional probability distributions of the projected data. 
The marginal distributions are aligned by estimating the coefficient matrix $\bA$, which minimizes: 
\begin{footnotesize}
	\begin{flalign}
		\min\limits_{\bA} \bigg|\bigg|\frac{1}{n_s}\sum\limits_{i=1}^{n_s}\bA^\top\bk_i - \frac{1}{n_t}\sum\limits_{j=n_s+1}^{n}\bA^\top\bk_j \bigg|\bigg|_{\cH}^2 = \tr(\bA^\top\bK\bM_0\bK^\top\bA).
		\label{Eq:JDA1}
	\end{flalign}
\end{footnotesize}
\noindent $\bM_0$, is the MMD matrix which given by, 
\begin{footnotesize}
  \begin{flalign}
    (\bM_0)_{ij} &= 
		\begin{cases}
				\frac{1}{n_sn_s},& \bx_i, \bx_j \in \cD_s\\
				\frac{1}{n_tn_t},& \bx_i, \bx_j \in \cD_t\\
				\frac{-1}{n_sn_t},& \text{otherwise},\\
		\end{cases}
  \end{flalign}
\end{footnotesize}
\noindent
Likewise, the conditional distribution difference can also be minimized by introducing matrices $M_c$, with $c = 1,\ldots,C$, defined as,
\begin{footnotesize}
	\begin{flalign}
    (\bM_c)_{ij} &= 
		\begin{cases}
				\frac{1}{n_s^{(c)}n_s^{(c)}},& \bx_i, \bx_j \in \cD_s^{(c)}\\
				\frac{1}{n_t^{(c)}n_t^{(c)}},& \bx_i, \bx_j \in \cD_t^{(c)}\\
				\frac{-1}{n_s^{(c)}n_t^{(c)}},& \begin{cases} 
																						\bx_i \in \cD_s^{(c)}, \bx_j \in \cD_t^{(c)}\\
																						\bx_j \in \cD_s^{(c)}, \bx_i \in \cD_t^{(c)}\\
																				\end{cases}\\
				0, & \text{otherwise}.
		\end{cases}
  \end{flalign}
\end{footnotesize}
\noindent Here, $\cD_s$ and $\cD_t$ are the sets of source and target data points respectively. 
$\cD_s^{(c)}$ is the subset of source data points with class label $c$ and $n_s^{(c)} = |\cD_s^{(c)}|$. 
Similarly, $\cD_t^{(c)}$ is the subset of target data points with class label $c$ and $n_t^{(c)} = |\cD_t^{(c)}|$. 
Since the target labels being unknown, we use predicted target labels to determine $\cD_t^{(c)}$. 
We initialize the target labels using a classifier trained on the source data and refine the labels over iterations. 
Combining both the marginal and conditional distribution terms leads us to the JDA model, which is given by,
\begin{footnotesize}
	\begin{flalign}
		\min\limits_{\bA} \sum\limits_{c=0}^{C}\tr(\bA^\top\bK\bM_c\bK^\top\bA).
		\label{Eq:JDA}
	\end{flalign}
\end{footnotesize}

\subsection{Similarity Based Embedding}
In addition to domain alignment, the NET algorithm ensures that the projected data $\bZ$, is classification friendly (easily classifiable). 
To this end we introduce laplacian eigenmaps in order to cluster datapoints based on class label similarity. 
The $(n\times n)$ adjacency matrix $\bW$, captures the similarity relationships between datapoints, where, 
\begin{footnotesize}
  \begin{flalign}
    \bW_{ij} \coloneqq
			\begin{cases}
				1 &~~~ y_i^s=y_j^s ~ \text{or} ~ i=j\\
				0 & y_i^s\neq y_j^s ~\text{or labels unknown.}
			\end{cases}
			\label{Eq:Adj}
  \end{flalign}
\end{footnotesize}
To ensure that the projected data is clustered based on data similarity, we minimize the sum of squared distances between data points weighted by the adjacency matrix. 
This can be expressed as a minimization problem, 
\begin{footnotesize}
  \begin{flalign}
    \min_{\bZ} \frac{1}{2}\sum_{ij}\bigg|\bigg|\frac{\bz_i}{\sqrt{d_i}} - \frac{\bz_j}{\sqrt{d_j}}\bigg|\bigg|^2\bW_{ij} = \min_{\bA}\tr(\bA^\top\bK\bL\bK^\top\bA).
		\label{Eq:Lapl}
  \end{flalign}
\end{footnotesize}
\noindent Here, $d_i = \sum_k\bW_{ik}$ and $d_j = \sum_k\bW_{jk}$. 
They form the diagonal entries of $\bD$, the $(n \times n)$ diagonal matrix. 
\begin{footnotesize}$||\bz_i/\sqrt{d_i} - \bz_j/\sqrt{d_j}||^2$\end{footnotesize}, is the squared normalized distance between the projected data points $\bz_i$ and $\bz_j$, which get clustered together when $\bW_{ij} = 1$, (as they belong to the same category). 
The normalized distance is a more robust clustering measure as compared to the standard Euclidean distance $||\bz_i - \bz_j||^2$, \cite{chung1997spectral}. 
Substituting \begin{footnotesize}$\bZ = \bA^\top\bK$\end{footnotesize}, yields the trace term, where $\bL$, denotes the symmetric positive semi-definite graph laplacian matrix with \begin{footnotesize}$\bL \coloneqq \bI-\bD^{-1/2}\bW\bD^{-1/2}$\end{footnotesize}, and $\bI$ is an identity matrix. 

\subsection{Optimization Problem}
\label{Sec:OptProb}
To arrive at the optimization problem, we consider the nonlinear projection in Equation (\ref{Eq:kpca}), the joint distribution alignment in Equation (\ref{Eq:JDA}) and the similarity based embedding in Equation (\ref{Eq:Lapl}). 
Maximizing Equation (\ref{Eq:kpca}) and minimizing Equations (\ref{Eq:JDA}) and (\ref{Eq:Lapl}) can also be achieved by maintaining Equation (\ref{Eq:kpca}) constant and minimizing Equations (\ref{Eq:JDA}) and (\ref{Eq:Lapl}). 
Minimizing the similarity embedding in Equation (\ref{Eq:Lapl}) can result in the projected vectors being embedded in a low dimensional subspace. 
In order to maintain the subspace dimensionality, we introduce a new constraint in place of Equation (\ref{Eq:kpca}). 
The optimization problem for NET is obtained by minimizing Equations (\ref{Eq:JDA}) and (\ref{Eq:Lapl}). 
The goal is to estimate the $(n\times k)$ projection matrix, $\bA$. 
Along with regularization and the dimensionality constraint, we get, 
\begin{footnotesize}
	\begin{flalign}
		\min_{\bA^\top\bK\bD\bK^\top\bA = \bI} &\alpha.\tr(\bA^\top\bK\sum_{c=0}^C\bM_c\bK^\top\bA) \notag\\
		&+ \beta.\tr(\bA^\top\bK \bL\bK^\top\bA) + \gamma||\bA||_F^2.
		\label{Eq:Opt}
	\end{flalign}
\end{footnotesize} 
The first term controls the domain alignment and is weighted by $\alpha$. 
The second term ensures similarity based embedding and is weighted by $\beta$. 
The third term is the regularization (Frobenius norm) that ensures a smooth projection matrix $\bA$ and it is weighted by $\gamma$. 
The constraint on $\bA$  (in place of \begin{footnotesize}$\bA^\top\bK\bH\bK^\top\bA = \bI$\end{footnotesize}), prevents the projection from collapsing onto a subspace with dimensionality less than $k$, \cite{belkin2003laplacian}. 
We solve Equation (\ref{Eq:Opt}) by forming the Lagrangian,
\begin{footnotesize}
	\begin{flalign}
		L(\bA, \mathbf{\Lambda)} = & \alpha.\tr\big(\bA^\top\bK \sum_{c=0}^C\bM_c\bK^\top\bA\big) + \beta.\tr(\bA^\top\bK \bL\bK^\top\bA) \notag  \\
															 &+ \gamma||\bA||_F^2 + \tr((\bI - \bA^\top\bK\bD\bK^\top\bA)\mathbf{\Lambda}),
		\label{Eq:Lagran}
	\end{flalign}
\end{footnotesize}
\noindent where the Lagrangian constants are represented by the diagonal matrix $\mathbf{\Lambda} = \textit{diag}(\lambda_1, \ldots, \lambda_k)$.  
Setting the derivative $\frac{\partial L}{\partial \bA} = 0$, yields the generalized eigen-value problem, 
\begin{footnotesize}
	\begin{flalign}
		\Big(\alpha\bK \sum_{c=0}^C\bM_c \bK^\top + \beta\bK \bL\bK^\top + \gamma\bI\Big)\bA = \bK\bD\bK^\top\bA\mathbf{\Lambda}.
		\label{Eq:GenEigen}
	\end{flalign}
\end{footnotesize}
\noindent The solution $\bA$ in (\ref{Eq:Opt}) are the $k$-smallest eigen-vectors of Equation (\ref{Eq:GenEigen}). 
The projected data points are then given by \begin{footnotesize}$\bZ = \bA^\top\bK$\end{footnotesize}. 
The NET algorithm is outlined in Algorithm \ref{Algo:NET}. 
\begin{algorithm}[t]
	\scriptsize
	\caption{Nonlinear Embedding Transform}\label{Algo:NET}
	\begin{algorithmic}[1]
	\Require $\bX$, $Y_S$, constants $\alpha, \beta$, regularization $\gamma$ and projection dimension $k$. 
	\Ensure Projection matrix $\bA$, projected data $\bZ$. 
	\State Compute kernel matrix $\bK$, for predefined kernel $k(.,.)$
	\State Define the adjacency matrix $\bW$ (Eq. (\ref{Eq:Adj})) 
	\State Compute $\bD = \text{diag}(d_1,\ldots,d_n)$, where $d_i = \sum_j\bW_{ij}$
	\State Compute normalized graph laplacian $\bL = \bI-\bD^{-1/2}\bW\bD^{-1/2}$
	\State Solve Eq (\ref{Eq:GenEigen}) and select $k$ smallest eigen-vectors as columns of $\bA$
	\State Estimate $\bZ \leftarrow \bA^\top \bK$
	\State Train a classifier with modified data $\{[\bz_1, \ldots, \bz_{n_s}], Y_S\}$
	\end{algorithmic}
\end{algorithm}

\subsection{Model Selection}
\label{Sec:ModSel}
In unsupervised domain adaptation the target labels are treated as unknown. 
Current domain adaptation methods that need to validate the optimum parameters for their models, inherently assume the availability of target labels \cite{long2013transfer}, \cite{long2014transfer}. 
However, in the case of real world applications, when target labels are not available, it is difficult to verify if the model parameters are optimal. 
In the case of the NET model, we have 4 parameters $(\alpha, \beta, \gamma, k)$, that we want to pre-determine. 
We introduce a technique using Kernel Mean Matching (KMM) to sample the source data to create a validation set. 
KMM has been used to weight source data points in order to reduce the distribution difference between the source and target data \cite{fernando2013unsupervised}, \cite{gong2013connecting}. 
Source data points with large weights have a similar marginal distribution to the target data. 
These data points are chosen to form the validation set. 
The KMM estimates the weights $w_i$, $i = 1, \ldots, n_s$, by minimizing \begin{footnotesize}$\big|\big|\frac{1}{n_s}\sum_{i=1}^{n_s}w_i\phi(\bx_i^s) -  \frac{1}{n_t}\sum_{j=1}^{n_t}\phi(\bx_j^t)\big|\big|_\mathcal{H}^2$\end{footnotesize}. 
In order to simplify, we define $\kappa_i := \frac{n_s}{n_t}\sum_{j=1}^{n_t}k(\bx_i^s, \bx_j^t)$, $i = 1,\ldots,n_s$ and \begin{footnotesize}$\bK_{S_{ij}} = k(\bx_i^s, \bx_j^s)$\end{footnotesize}. 
The minimization is then represented as a quadratic programming problem, 
\begin{footnotesize}
	\begin{flalign}
		\min_{\bw} =& \frac{1}{2} \bw^\top\bK_S\bw - \mathbf{\kappa}^\top\bw, \notag \\
		&\text{s.t.}~ w_i\in[0, B], ~\bigg|\sum_{i=1}^{n_s}w_i - n_s \bigg|\leq n_s\epsilon.
		\label{Eq:KMM}
	\end{flalign}
\end{footnotesize}
The first constraint limits the scope of discrepancy between source and target distributions, with $B\rightarrow 1$, leading to an unweighted solution. 
The second constraint ensures the measure $w(x)P_S(x)$, is a probability distribution \cite{gretton2009covariate}. 
In our experiments, we select 10\% of the source data with the largest weights to create the validation set. 
We fine tune the values of $(\alpha, \beta, \gamma, k)$, using the validation set. 
For fixed values of $(\alpha, \beta, \gamma, k)$, the NET model is trained using the source data (without the validation set) and target data. 
The model is tested on the validation set to estimate parameters yielding highest classification accuracies. 

\section{Experiments}
In this section, we evaluate the NET algorithm and the model selection proposition across multiple image classification datasets and several existing procedures for unsupervised domain adaptation. 

\subsection{Datasets}
We conduct our experiments across 7 different datasets. Their characteristics are outlined in Table (\ref{Tab:Datasets}). 

\noindent\textbf{\textit{MNIST-USPS} datasets}: These are popular handwritten digit recognition datasets. 
Here, the digit images are subsampled to  $16 \times 16$ pixels. 
Based on \cite{long2014transfer}, we consider two domains \texttt{MNIST} (2,000 images from MNIST) and \texttt{USPS} (1,800 images from USPS). 

\noindent\textbf{\textit{CKPlus-MMI} datasets}: The CKPlus \cite{lucey2010extended}, and MMI \cite{pantic2005web} are popular Facial Expression recognition datasets. 
They contain videos of facial expressions. 
We choose 6 categories of facial expression, viz., \emph{anger, disgust, fear, happy, sad, surprise}. 
We create two domains, \texttt{CKPlus} and \texttt{MMI}, by selecting video frames with the most intense expressions. 
We use a pre-trained deep convolutional neural network (CNN), to extract features from these images. 
In our experiments, we use the VGG-F model \cite{Chatfield14}, trained on the popular ImageNet object recognition dataset. 
The VGG-F network is similar in architecture to the popular AlexNet \cite{krizhevsky2012imagenet}. 
We extract the 4096-dimensional features that are fed into the fully-connected $fc8$ layer. 
We apply PCA on the combined source and target data to reduce the dimension to 500 and use these features across all the experiments. 

\noindent\textbf{\textit{COIL20} dataset}: It is an object recognition dataset consisting of 20 categories with two domains, \texttt{COIL1} and \texttt{COIL2}. 
The domains consist of images of objects captured from views that are 5 degrees apart. The images are  $32 \times 32$ pixels with gray scale values \cite{long2013transfer}. 

\noindent\textbf{\textit{PIE} dataset}: The ``Pose, Illumination and Expression'' (PIE) dataset consists of face images ( $32 \times 32$ pixels) of 68 individuals. 
The images were captured with different head-pose, illumination and expression. 
Similar to \cite{long2013transfer}, we select 5 subsets with differing head-pose to create 5 domains, namely, \texttt{P05} (C05, left pose), \texttt{P07} (C07, upward pose), \texttt{P09} (C09, downward pose), \texttt{P27} (C27, frontal pose) and \texttt{P29} (C29, right pose). 

\noindent\textbf{\textit{Office-Caltech} dataset}: This is currently the most popular benchmark dataset for object recognition in the domain adaptation computer vision community. 
The dataset consists of images of everyday objects. 
It consists of 4 domains; \texttt{Amazon}, \texttt{Dslr} and \texttt{Webcam} from the \textit{Office} dataset and \texttt{Caltech} domain from the \textit{Caltech-256} dataset.
The \texttt{Amazon} domain has images downloaded from the www.amazon.com website. 
The \texttt{Dslr} and \texttt{Webcam} domains have images captured using a DSLR camera and a webcam respectively. 
The \texttt{Caltech} domain is a subset of the Caltech-256 dataset that was created by selecting categories common with the \textit{Office} dataset.  
The \textit{Office-Caltech} dataset has 10 categories of objects and a total of 2533 images (data points). 
We experiment with two kinds of features for the \textit{Office-Caltech} dataset; (i) 800-dimensional SURF features \cite{gong2012geodesic}, (ii) Deep features. 
The deep features are extracted using a pre-trained network similar to the \textit{CKPlus-MMI} datasets. 

\begin{table}[!ht]
\scriptsize
\centering
\caption{Statistics for the benchmark datasets}
\label{Tab:Datasets}
\resizebox{\linewidth}{!}{%
\begin{tabular}{|c|c|c|c|c|c|}
\hline
\textbf{Dataset} & \textbf{Type} & \textbf{\#Samples} & \textbf{\#Features} & \textbf{\#Classes} & \textbf{Subsets}\\\hline\hline
MNIST & Digit & 2,000 & 256 & 10 & MNIST \\\hline
USPS & Digit & 1,800 & 256 & 10 & USPS \\\hline
CKPlus & Face Exp. & 1,496 & 4096 & 6 & CKPlus \\\hline
MMI & Face Exp. & 1,565 & 4096 & 6 & MMI \\\hline
COIL20 & Object & 1,440 & 1,024 & 20 & COIL1, COIL2 \\\hline
PIE & Face & 11,554 & 1,024 & 68 & P05, ..., P29 \\\hline
Ofc-Cal SURF & Object & 2,533 & 800 & 10 & A, C, W, D \\\hline
Ofc-Cal Deep & Object & 2,505 & 4096 & 10 & A, C, W, D \\\hline
\end{tabular}
}
\end{table}

\subsection{Existing Baselines}
We compare the NET algorithm with the following baseline and state-of-the-art methods. 
\begin{table}[!ht]
\scriptsize
\centering
\caption{Baseline methods that are compared with the NET.}
\label{Tab:Baseline}
\resizebox{\linewidth}{!}{%
\begin{tabular}{|c|p{0.7\linewidth}|}
\hline
\textbf{Method} & \textbf{Reference} \\\hline\hline
SA & Subspace Alignment \cite{fernando2013unsupervised} \\\hline
CA & Correlation Alignment \cite{sun2015return} \\\hline
GFK & Geodesic Flow Kernel \cite{gong2012geodesic} \\\hline
TCA & Transfer Component Analysis \cite{pan2011domain} \\\hline
TJM & Transfer Joint Matching \cite{long2014transfer} \\\hline
JDA & Joint Distribution Adaptation \cite{long2013transfer}  \\\hline
\end{tabular}
}
\end{table} 
Like NET, the TCA, TJM and JDA are all spectral methods. 
While all the four algorithms use MMD to align the source and target datasets, the NET, in addition, uses nonlinear embedding for classification enhancement. 
TCA, TJM and JDA, solve for \begin{footnotesize}$\bA$\end{footnotesize} in a setting similar to Equation (\ref{Eq:GenEigen}). 
However, unlike NET, they do not have the similarity based embedding term and $\alpha = 1$, is fixed in all the three algorithms. 
Therefore, these models have only 2 free parameters $(\gamma ~\text{and} ~k)$, that need to be pre-determined in contrast to NET, which has 4 parameters, $(\alpha, \beta, \gamma, k)$. 
Since TCA, TJM and JDA, are all quite similar to each other, for the sake of brevity, we evaluate model selection (estimating optimal model parameters) using only JDA and NET. 
The SA, CA and GFK algorithms, do not have any critical free model parameters that need to be pre-determined. 

In our experiments, $\text{NET}_v$ is a special case of the NET, where model parameters $(\alpha, \beta, \gamma, k)$, have been determined using a validation set derived from Equation (\ref{Eq:KMM}). 
Similarly, $\text{JDA}_v$ is a special case of JDA, where $(\gamma, k)$, have been determined using a validation set derived from Equation (\ref{Eq:KMM}). 
In order to ascertain the optimal nature of the parameters determined with a source-based validation set, we estimate the best model parameters using the target data (with labels) as a validation set. 
These results are represented by $\text{NET}$ in the figures and tables. 
The results for the rest of the algorithms (SA, CA, GFK, TCA, TJM and JDA), are obtained with the parameter settings described in their respective works. 

\subsection{Experimental Details}
For fair comparison with existing methods, we follow the same experimental protocol as in \cite{gong2012geodesic,long2014transfer}. 
We conduct 50 different domain adaptation experiments with the previously mentioned datasets. 
Each of these is an unsupervised domain adaptation experiment with one source domain (data points and labels) and one target domain (data points only). 
When estimating $\bM_c$, we choose 10 iterations to converge to the predicted test/validation labels. 
Wherever necessary, we use a Gaussian kernel for $k(.,.)$, with a standard width equal to the median of the squared distances over the dataset. 
We train a 1-Nearest Neighbor (NN) classifier using the projected source data and test on the projected target data for all the experiments. 
We choose a NN classifier as in \cite{gong2012geodesic,long2014transfer}, since it does not require tuning of cross-validation parameters. 
The accuracies reflect the percentage of correctly classified target data points. 
\begin{table}[t]
\centering
\caption{Recognition accuracies (\%) for domain adaptation experiments on the digit and face datasets. \{\texttt{MNIST}(M), \texttt{USPS}(U), \texttt{CKPlus}(CK), \texttt{MMI}(MM), \texttt{COIL1}(C1) and \texttt{COIL2}(C2). M$\rightarrow$U implies M is source domain and U is target domain. The best and second best results in every experiment (row) are in \textbf{bold} and \textit{italic} respectively. The shaded columns indicate accuracies obtained with model selection.}
\label{Tab:DigitCoilFacePie}
\resizebox{\linewidth}{!}{%
\begin{tabular}{|c|c|c|c|c|c|c|g|c|g|}
\hline
 \textbf{Expt.} &  \textbf{SA} &  \textbf{CA} &  \textbf{GFK} & \textbf{TCA} & \textbf{TJM} & \textbf{JDA} &  $\textbf{JDA}_v$ &  \textbf{NET} & $\textbf{NET}_v$ \\\hline\hline
 M$\rightarrow$U  & 67.39  & 59.33  & 66.06  & 60.17  & 64.94  & 67.28  & 71.94  & \textbf{75.39}  & \textit{72.72} \\\hline
 U$\rightarrow$M  & 51.85  & 50.80  & 47.40  & 39.85  & 52.80  & 59.65  & 59.65  & \textbf{62.60}  & \textit{61.35} \\\hline
 C1$\rightarrow$C2  & 85.97  & 84.72  & 85.00  & 90.14  & 91.67  & 92.64  & \textbf{95.28}  & \textit{93.89}  & 90.42 \\\hline
 C2$\rightarrow$C1  & 84.17  & 82.78  & 84.72  & 88.33  & 89.86  & \textit{93.75}  & \textbf{93.89}  & 92.64  & 88.61 \\\hline
 CK$\rightarrow$MM  & \textit{31.12}  & \textbf{31.89}  & 28.75  & 32.72  & 30.35  & 29.78  & 25.82  & 29.97  & 30.54 \\\hline
 MM$\rightarrow$CK  & 39.75  & 37.74  & 37.94  & 31.33  & \textit{40.62}  & 28.39  & 26.79  & \textbf{45.83}  & 40.08 \\\hline
 P05$\rightarrow$P07  & 26.64  & 40.33  & 26.21  & 40.76  & 10.80  & 58.81  & \textit{77.53}  & \textbf{77.84}  & 69.00 \\\hline
 P05$\rightarrow$P09  & 27.39  & 41.97  & 27.27  & 41.79  & 7.29  & 54.23  & \textit{66.42}  & \textbf{70.96}  & 57.41 \\\hline
 P05$\rightarrow$P27  & 30.28  & 55.36  & 31.15  & 59.60  & 15.14  & 84.50  & \textit{90.78}  & \textbf{91.86}  & 84.68 \\\hline
 P05$\rightarrow$P29  & 19.24  & 29.04  & 17.59  & 29.29  & 4.72  & 49.75  & \textbf{52.70}  & \textit{52.08}  & 45.40 \\\hline
 P07$\rightarrow$P05  & 25.42  & 41.51  & 25.27  & 41.78  & 16.63  & 57.62  & \textbf{74.70}  & \textit{74.55}  & 57.92 \\\hline
 P07$\rightarrow$P09  & 47.24  & 53.43  & 47.37  & 51.47  & 21.69  & 62.93  & \textbf{79.66}  & \textit{77.08}  & 54.60 \\\hline
 P07$\rightarrow$P27  & 53.47  & 63.77  & 54.22  & 64.73  & 26.04  & 75.82  & 81.14  & \textit{83.84}  & \textbf{86.09} \\\hline
 P07$\rightarrow$P29  & 26.84  & 35.72  & 27.02  & 33.70  & 10.36  & 39.89  & \textit{63.73}  & \textbf{69.24}  & 47.30 \\\hline
 P09$\rightarrow$P05  & 23.26  & 35.47  & 21.88  & 34.69  & 14.98  & 50.96  & \textbf{77.16}  & \textit{73.98}  & 68.67 \\\hline
 P09$\rightarrow$P07  & 41.87  & 47.08  & 43.09  & 47.70  & 27.26  & 57.95  & \textit{78.39}  & \textbf{79.01}  & 67.34 \\\hline
 P09$\rightarrow$P27  & 44.97  & 53.71  & 46.38  & 56.23  & 27.55  & 68.45  & \textit{84.92}  & 83.48  & \textbf{87.47} \\\hline
 P09$\rightarrow$P29  & 28.13  & 34.68  & 26.84  & 33.09  & 8.15  & 39.95  & 65.93  & \textbf{70.04}  & \textit{67.65} \\\hline
 P27$\rightarrow$P05  & 35.62  & 51.17  & 34.27  & 55.61  & 25.96  & 80.58  & \textit{92.83}  & \textbf{93.07}  & 92.44 \\\hline
 P27$\rightarrow$P07  & 63.66  & 66.05  & 62.92  & 67.83  & 28.73  & 82.63  & \textit{90.18}  & 89.99  & \textbf{93.68} \\\hline
 P27$\rightarrow$P09  & 72.24  & 73.96  & 73.35  & 75.86  & 38.36  & 87.25  & \textit{90.14}  & 89.71  & \textbf{90.20} \\\hline
 P27$\rightarrow$P29  & 36.03  & 40.50  & 37.38  & 40.26  & 7.97  & 54.66  & 72.18  & \textit{76.84}  & \textbf{79.53} \\\hline
 P29$\rightarrow$P05  & 23.05  & 26.89  & 20.35  & 27.01  & 9.54  & 46.46  & \textit{60.20}  & \textbf{67.32}  & 52.67 \\\hline
 P29$\rightarrow$P07  & 26.03  & 31.74  & 24.62  & 29.90  & 8.41  & 42.05  & \textbf{71.39}  & \textit{70.23}  & 57.52 \\\hline
 P29$\rightarrow$P09  & 27.76  & 31.92  & 28.49  & 29.90  & 6.68  & 53.31  & \textit{74.02}  & \textbf{74.63}  & 62.81 \\\hline
 P29$\rightarrow$P27  & 30.31  & 34.70  & 31.27  & 33.67  & 10.06  & 57.01  & \textit{76.66}  & 75.43  & \textbf{80.98} \\\hline\hline
 \textbf{Average}  & 41.14  & 47.55  & 40.65  & 47.59  & 26.79  & 60.63  & \textit{72.85}  & \textbf{74.67}  & 68.73 \\\hline
\end{tabular}
}
\end{table}

\begin{table*}[t]
\centering
\caption{Recognition accuracies (\%) for domain adaptation experiments on the \textit{Office-Caltech} dataset with SURF and Deep features. \{\texttt{Amazon}(A), \texttt{Webcam}(W), \texttt{Dslr}(D), \texttt{Caltech}(C)\}. A$\rightarrow$W implies A is source and W is target. The best and second best results in every experiment (row) are in \textbf{bold} and \textit{italic} respectively. The shaded columns indicate accuracies obtained with model selection.}
\label{Tab:OfcCal}
\resizebox{\textwidth}{!}{%
\begin{tabular}{|c|c|c|c|c|c|c|g|c|g||c|c|c|c|c|c|g|c|g|}
\hline
\multirow{2}{*}{\textbf{Expt.}} & \multicolumn{9}{c||}{\textbf{SURF Features}} & \multicolumn{9}{c|}{\textbf{Deep Features}}\\
\hhline{~------------------}
&  \textbf{SA} &  \textbf{CA} &  \textbf{GFK} &  \textbf{TCA} &  \textbf{TJM} &  \textbf{JDA} &  $\textbf{JDA}_v$ &  \textbf{NET} &  $\textbf{NET}_v$ &  \textbf{SA} &  \textbf{CA} &  \textbf{GFK} &  \textbf{TCA} &  \textbf{TJM} &  \textbf{JDA} &  $\textbf{JDA}_v$ &  \textbf{NET} & $\textbf{NET}_v$ \\\hline\hline
 C$\rightarrow$A   & 43.11  & 36.33  & 45.72  & 44.47  & \textbf{46.76}  & 44.78  & 45.41  & \textit{46.45}  & 46.24  & 88.82  & \textit{91.12}  & 90.60  & 89.13  & 91.01  & 90.07  & 89.34  & \textbf{92.48}  & 90.70 \\\hline
 D$\rightarrow$A   & 29.65  & 28.39  & 26.10  & 31.63  & 32.78  & 33.09  & 29.85  & \textbf{39.67}  & \textit{35.60}  & 84.33  & 86.63  & 88.40  & 88.19  & 88.72  & 91.22  & 90.18  & \textbf{91.54}  & \textit{91.43} \\\hline
 W$\rightarrow$A   & 32.36  & 31.42  & 27.77  & 29.44  & 29.96  & 32.78  & 29.33  & \textbf{41.65}  & \textit{39.46}  & 84.01  & 82.76  & 88.61  & 86.21  & 88.09  & 91.43  & 87.04  & \textbf{92.58}  & \textit{91.95} \\\hline 
 A$\rightarrow$C   & 38.56  & 33.84  & 39.27  & 39.89  & 39.45  & 39.36  & 39.27  & \textbf{43.54}  & \textit{43.10}  & 80.55  & \textit{82.47}  & 81.01  & 75.53  & 78.08  & 83.01  & 78.27  & \textbf{83.01}  & 82.28 \\\hline
 D$\rightarrow$C   & 31.88  & 29.56  & 30.45  & 30.99  & 31.43  & 31.52  & 31.08  & \textbf{35.71}  & \textit{34.11}  & 76.26  & 75.98  & 78.63  & 74.43  & 76.07  & 80.09  & 78.17  & \textit{82.10}  & \textbf{83.38} \\\hline
 W$\rightarrow$C   & 29.92  & 28.76  & 28.41  & 32.15  & 30.19  & 31.17  & 31.43  & \textbf{35.89}  & \textit{32.77}  & 78.90  & 74.98  & 76.80  & 76.71  & 79.18  & \textbf{82.74}  & 78.90  & \textit{82.56}  & 82.28 \\\hline
 A$\rightarrow$D   & 37.58  & 36.94  & 34.40  & 33.76  & \textbf{45.22}  & 39.49  & 31.85  & \textit{40.76}  & 36.31  & 82.17  & \textit{87.90}  & 82.80  & 82.17  & 87.26  & 89.81  & 77.07  & \textbf{91.08}  & 80.89 \\\hline
 C$\rightarrow$D   & 43.95  & 38.22  & 43.31  & 36.94  & 44.59  & \textit{45.22}  & 40.13  & \textbf{45.86}  & 36.31  & 80.89  & 82.80  & 77.07  & 75.80  & 82.80  & 89.17  & 80.25  & \textbf{92.36}  & \textit{90.45} \\\hline
 W$\rightarrow$D   & \textit{90.45}  & 85.35  & 82.17  & 85.35  & 89.17  & 89.17  & 88.53  & 89.81  & \textbf{91.72}  & \textbf{100.00}  & \textbf{100.00}  & \textbf{100.00}  & \textbf{100.00}  & \textbf{100.00}  & \textbf{100.00}  & \textbf{100.00}  & \textit{99.36}  & \textbf{100.00} \\\hline
 A$\rightarrow$W   & 37.29  & 31.19  & 41.70  & 33.90  & \textit{42.03}  & 37.97  & 38.98  & \textbf{44.41}  & 35.25  & 82.37  & 80.34  & 84.41  & 76.61  & 87.12  & 87.12  & 79.32  & \textbf{90.85}  & \textit{87.46} \\\hline
 C$\rightarrow$W   & 36.27  & 29.49  & 35.59  & 32.88  & 38.98  & \textit{41.69}  & 37.97  & \textbf{44.41}  & 33.56  & 77.29  & 79.32  & 78.64  & 78.31  & \textit{88.48}  & 85.76  & 77.97  & \textbf{90.85}  & 84.07 \\\hline
 D$\rightarrow$W   & 87.80  & 83.39  & 79.66  & 85.42  & 85.42  & \textit{89.49}  & 86.78  & 87.80  & \textbf{90.51}  & 98.98  & \textit{99.32}  & 98.31  & 97.97  & 98.31  & 98.98  & 98.98  & \textbf{99.66}  & \textbf{99.66} \\\hline
 \hline
 \textbf{Average}  & 44.90  & 41.07  & 42.88  & 43.07  & \textit{46.33}  & 46.31  & 44.22  & \textbf{49.66}  & 46.24  & 84.55  & 85.30  & 85.44  & 83.42  & 87.09  & \textit{89.12}  & 84.63  & \textbf{90.70}  & 88.71 \\\hline
\end{tabular}
}
\end{table*}

\begin{figure*}[t]
\centering
\subfloat[\scriptsize{\# bases $k$}]{
		\label{Fig:NETkStudy}
    \includegraphics[width=0.24\textwidth]{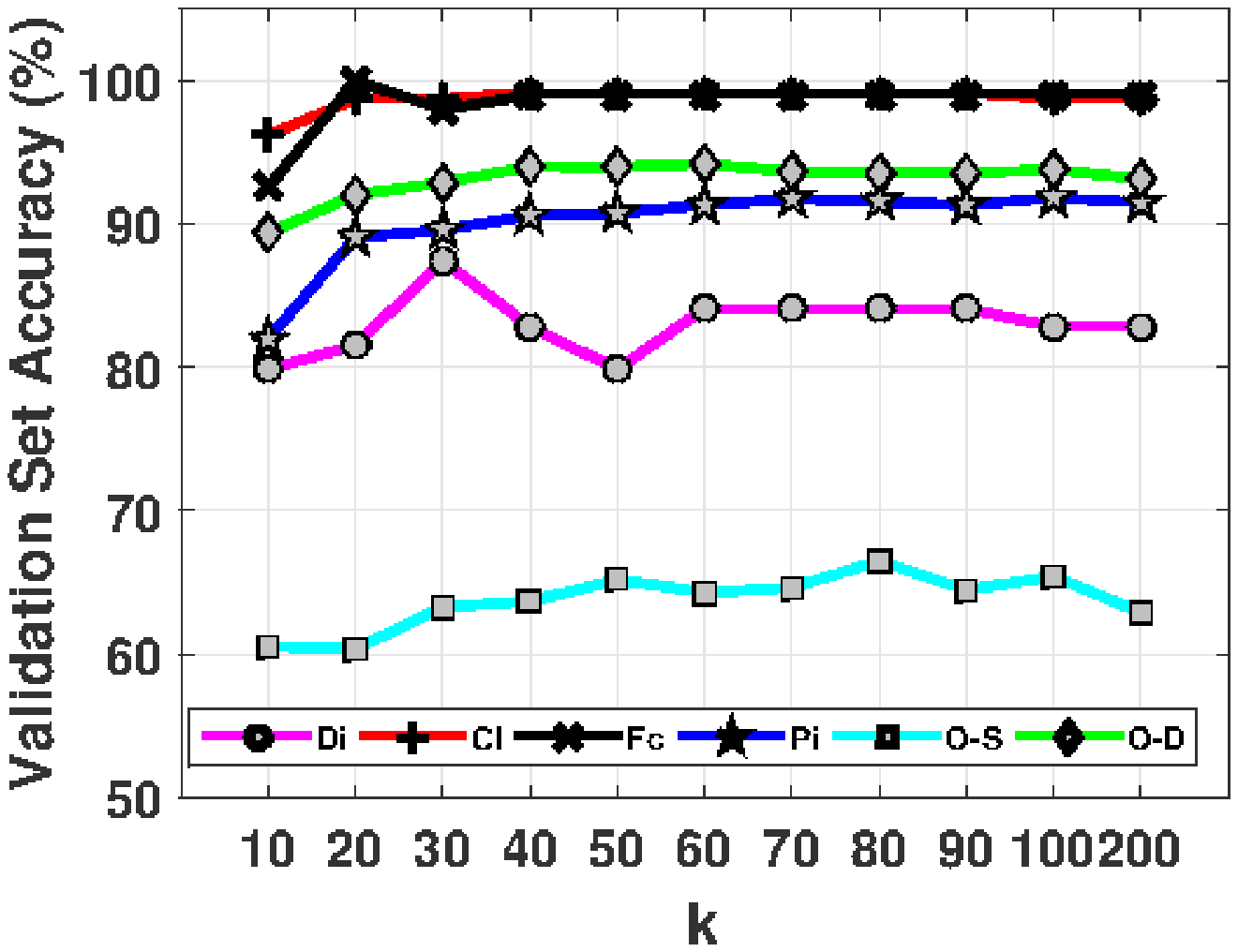}
}%
\hfill
\subfloat[\scriptsize{MMD weight $\alpha$}]{
		\label{Fig:NETaStudy}
    \includegraphics[width=0.24\textwidth]{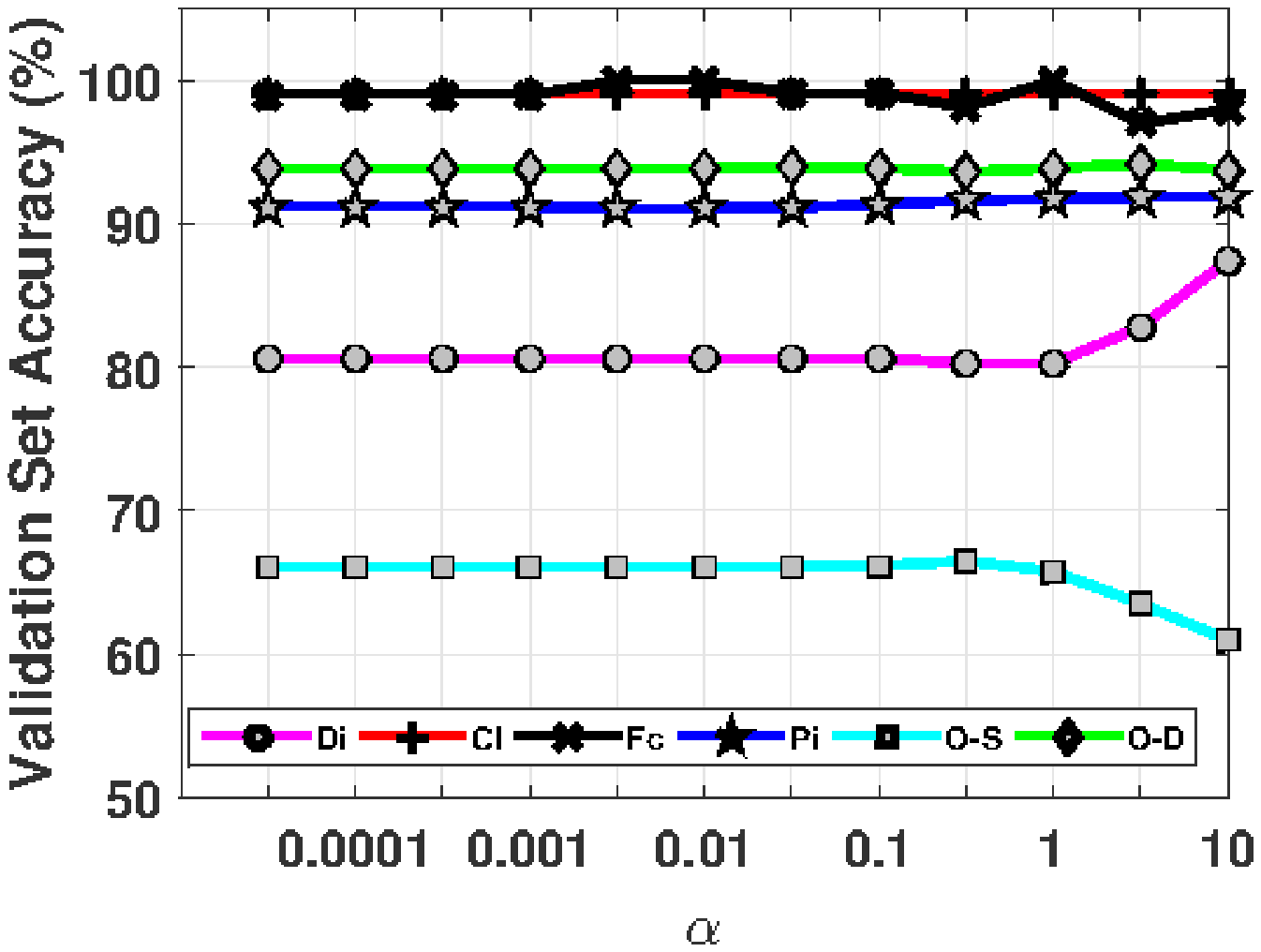}
}%
\hfill
\subfloat[\scriptsize{Embed weight $\beta$}]{
		\label{Fig:NETbStudy}
    \includegraphics[width=0.24\textwidth]{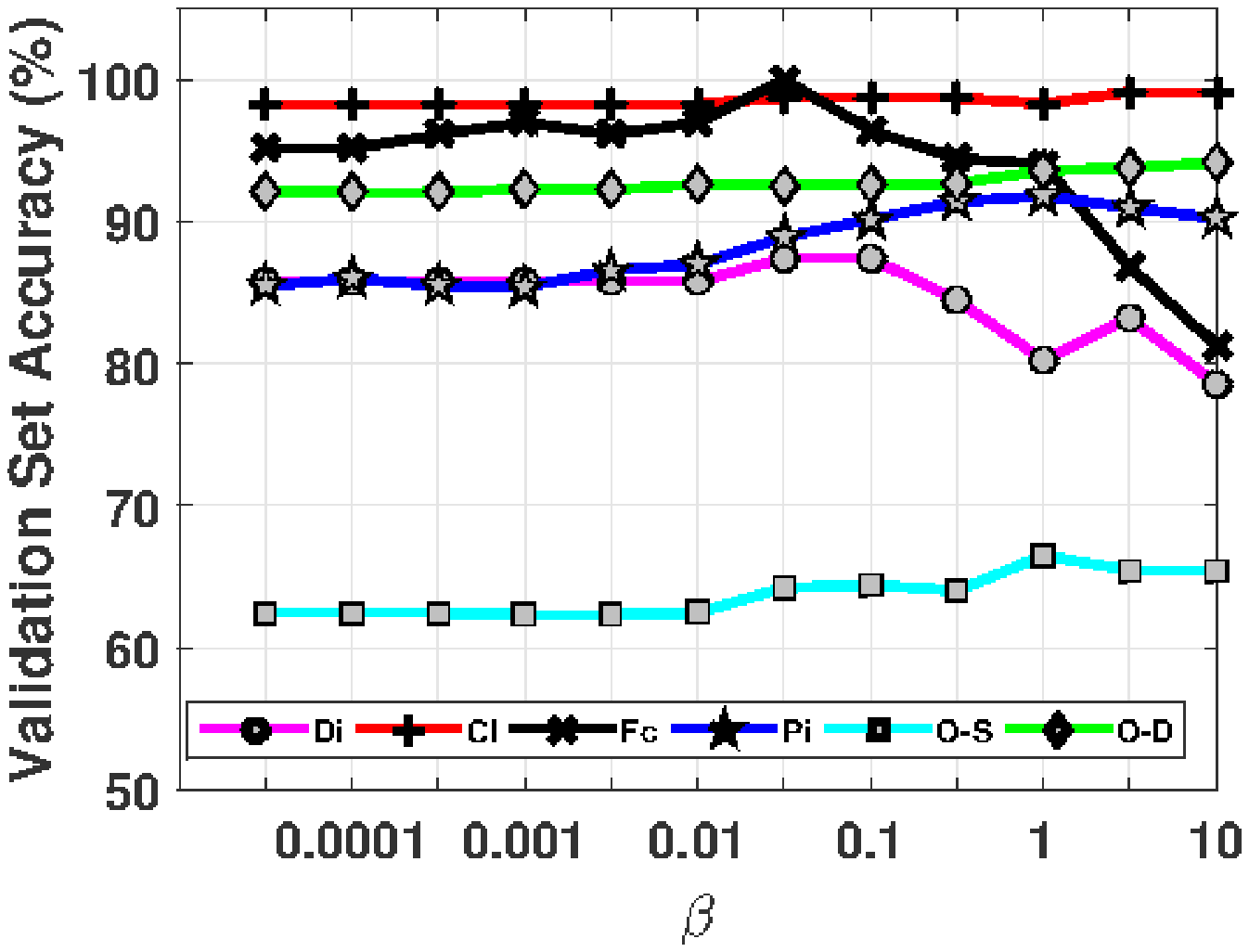}
}%
\hfill
\subfloat[\scriptsize{Regularization $\gamma$}]{
		\label{Fig:NETgStudy}
    \includegraphics[width=0.24\textwidth]{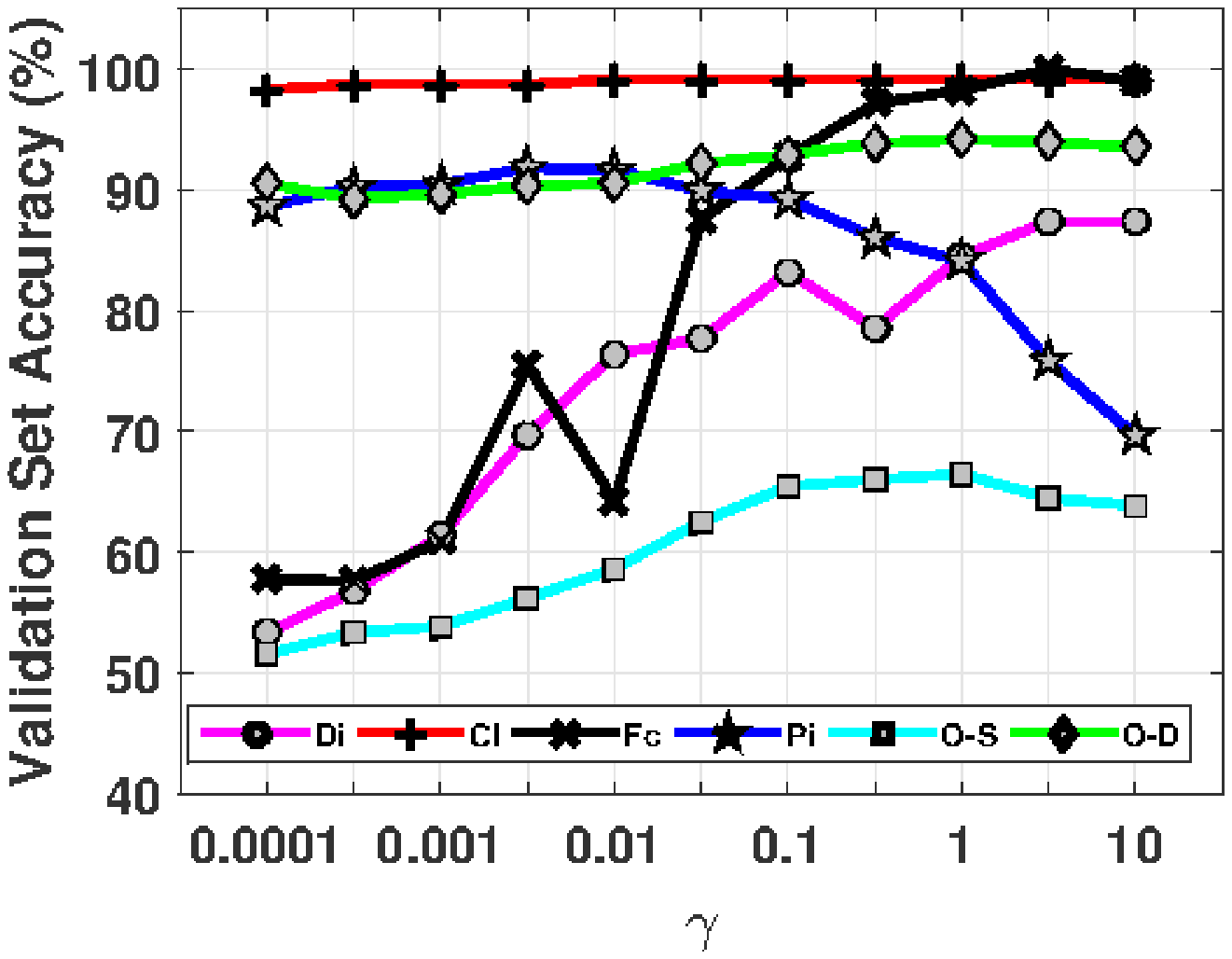}
}%
\caption{NET Validation Study. Each figure depicts the accuracies over the source-based validation set. When studying a parameter (say $k$), the remaining parameters $(\alpha, \beta, \gamma)$ are fixed at the optimum value. The legend is, Digit (Di), Coil (Cl), MMI$\&$CK+ Face (Fc), PIE (Pi), Office-Caltech SURF (O-S) and Office-Caltech Deep (O-D).}
\label{Fig:NETvStudy}
\end{figure*}

\subsection{Parameter Estimation Study}
Here we evaluate our model selection procedure. 
The NET algorithm has 4 parameters $(\alpha, \beta, \gamma, k)$, and the JDA has 2 parameters $(\gamma, k)$, that need to be pre-determined. 
To determine these parameters, we weight the source data points using Equation (\ref{Eq:KMM}) and select 10\% of the source data points with the largest weights. 
These source data points have a distribution similar to the target and they are used as a validation set to determine the optimal values for the model parameters $(\alpha, \beta, \gamma, k)$. 
The parameter space consists of $k \in \{10, 20, \ldots, 100, 200\}$ and $\alpha, \beta, \gamma$ from the set $\{0, 0.0001, 0.0005, 0.001, 0.005, 0.01, 0.05, 0.1, 0.5, 1, 5, 10\}$. 
For the sake of brevity, we present one set of parameters for every dataset, although in practice, a unique set of parameters can be evaluated for every domain adaptation experiment. 
Given a set of model parameters, we conduct the domain adaptation experiment using the entire source data (data and labels) and the target data (data only). 
The accuracies obtained are represented as shaded columns $\text{JDA}_v$ and $\text{NET}_v$ in Tables (\ref{Tab:DigitCoilFacePie}) and (\ref{Tab:OfcCal}). 

\begin{figure}[t]
\centering
\subfloat[\scriptsize{\# bases $k$}]{
		\label{Fig:JDAkStudy}
    \includegraphics[width=0.22\textwidth]{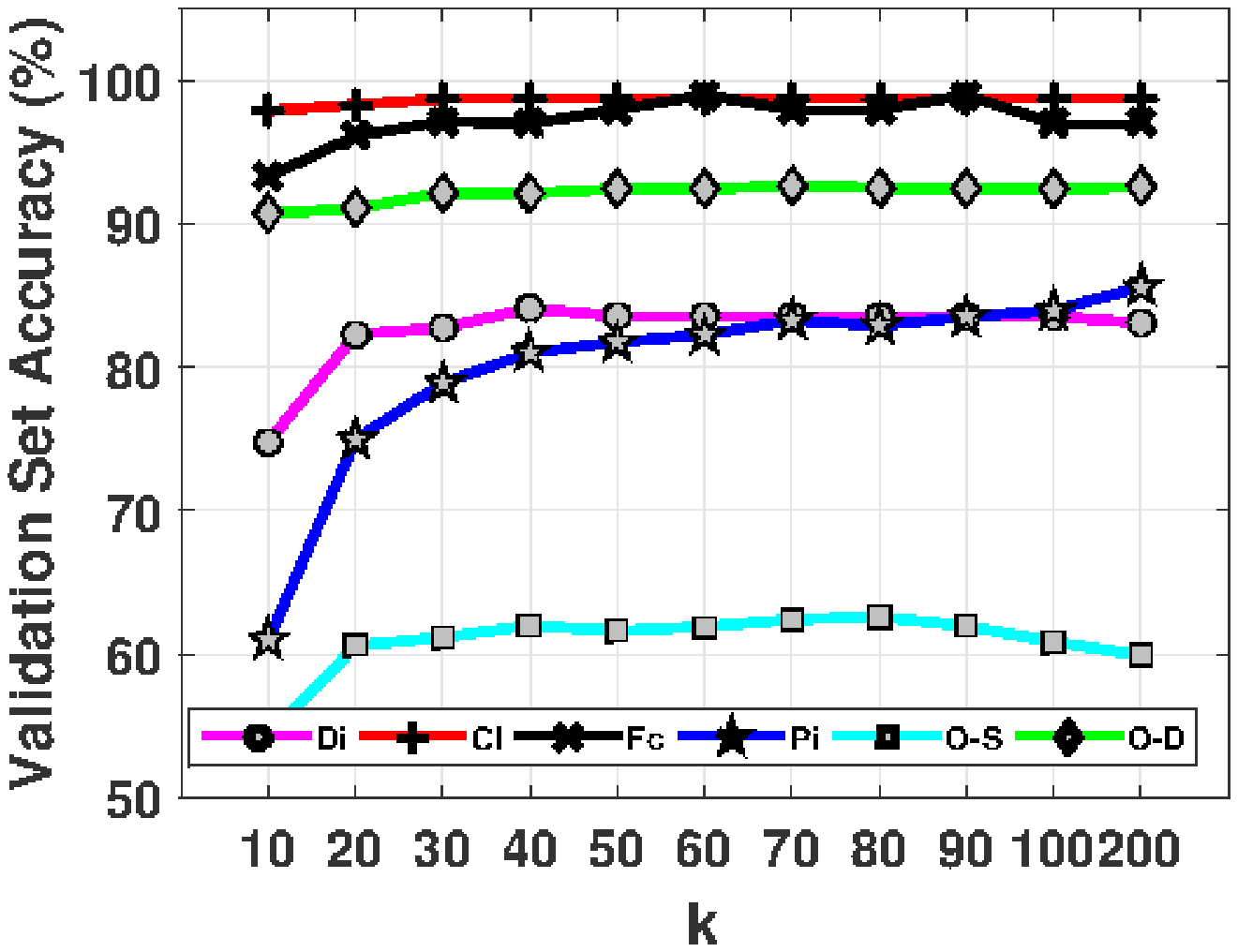}
}%
\hfill
\subfloat[\scriptsize{Regularization $\gamma$}]{
		\label{Fig:JDAgStudy}
    \includegraphics[width=0.22\textwidth]{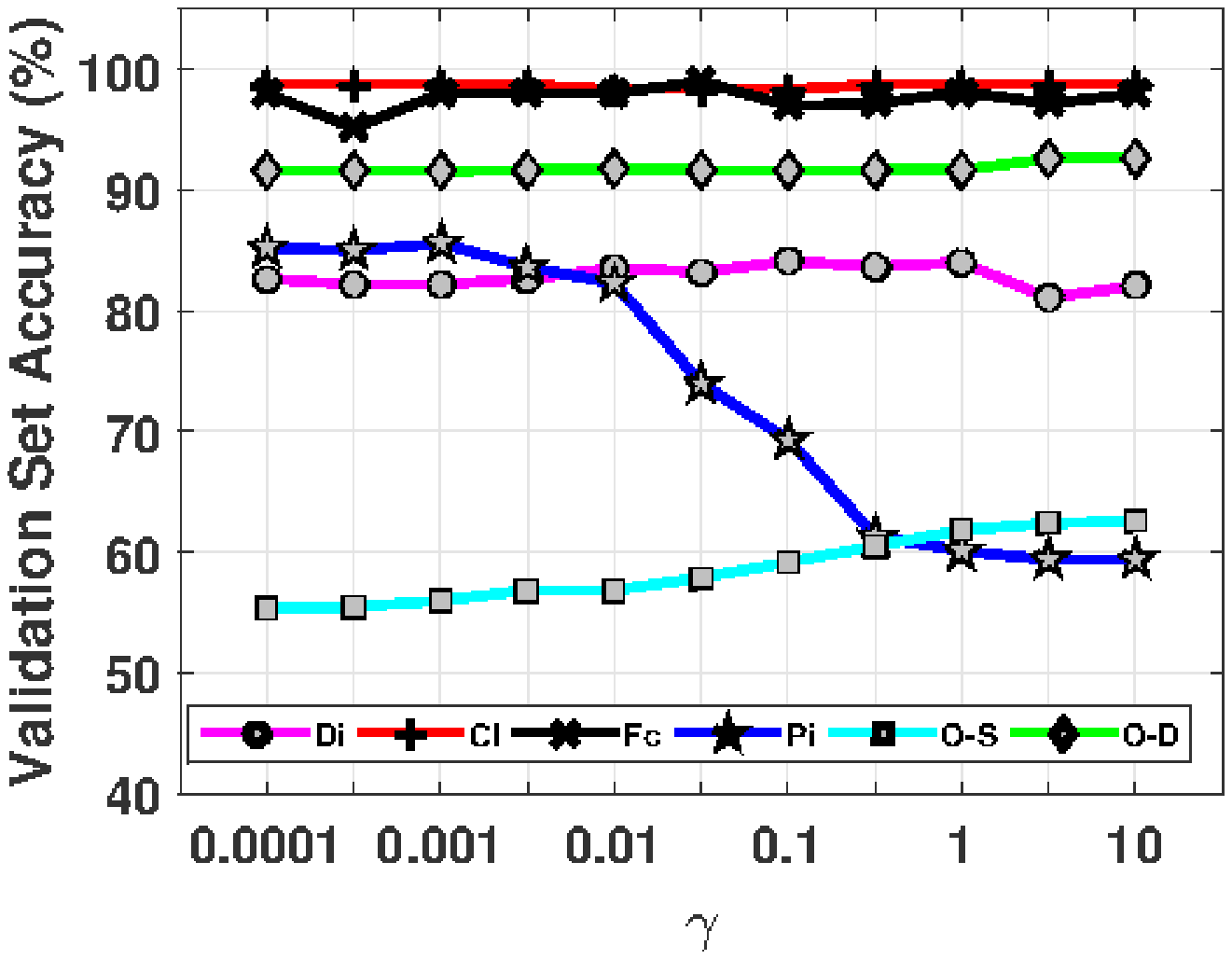}
}%
\caption{JDA Validation Study. Each figure depicts the accuracies over the source-based validation set. When studying a parameter (say $k$), the remaining parameter $\gamma$ is fixed at the optimum value. The legend is, Digit (Di), Coil (Cl), MMI$\&$CK+ Face (Fc), PIE (Pi), Office-Caltech SURF (O-S) and Office-Caltech Deep (O-D).}
\label{Fig:JDAvStudy}
\end{figure}

In order to evaluate the validity of our proposed model selection method, we also determine the parameters using the target data as a validation set.
These results are represented by the NET column in Tables (\ref{Tab:DigitCoilFacePie}) and (\ref{Tab:OfcCal}). 
Since the NET column values have been determined using the target data, they can be considered as the best accuracies for the NET model. 
The rest of the column values SA, CA, GFK, TCA, TJM and JDA, were estimated with model parameters suggested in their respective papers.  
The recognition accuracies for $\text{NET}_v$ is greater than that of the other domain adaptation methods and is nearly comparable to the NET. 
In Table (\ref{Tab:DigitCoilFacePie}), the $\text{JDA}_v$ has better performance than the JDA. 
This goes to show that a proper validation procedure does help select the best set of model parameters. 
It demonstrates that the proposed model selection procedure is a valid technique for evaluating an unsupervised domain adaptation algorithm in the absence of target data labels. 
Figures (\ref{Fig:NETvStudy}) and (\ref{Fig:JDAvStudy}), depict the variation of average validation set accuracies over the model parameters. 
Based on these curves, the optimal parameters are chosen for each of the datasets.

\subsection{NET Algorithm Evaluation}
The NET algorithm has been compared to existing unsupervised domain adaptation procedures across multiple datasets. 
The results of the NET algorithm are depicted under the NET column in Tables (\ref{Tab:DigitCoilFacePie}) and (\ref{Tab:OfcCal}). 
The parameters used to obtain these results are depicted in Table (\ref{Tab:NETParam}). 
The accuracies obtained with the NET algorithm are consistently better than any of the other spectral methods (TCA, TJM and JDA). 
NET also consistently performs better compared to non-spectral methods like SA, CA and GFK. 
\begin{table}[!ht]
\centering
\caption{Parameters used for the NET model.}
\label{Tab:NETParam}
\resizebox{0.7\linewidth}{!}{%
\tiny
\begin{tabular}{|c|c|c|c|c|}
\hline
\textbf{Dataset} & \textbf{$\alpha$} & \textbf{$\beta$} & \textbf{$\gamma$} & \textbf{$k$} \\\hline\hline
MNIST $\&$ USPS & 1.0 & 0.01 & 1.0 & 20 \\\hline
MMI $\&$ CK+ & 0.01 & 0.01 & 1.0 & 20 \\\hline
COIL & 1.0 & 1.0 & 1.0 & 60 \\\hline
PIE & 10.0 & 0.001 & 0.005 & 200 \\\hline
Ofc-SURF & 1.0 & 1.0 & 1.0 & 20 \\\hline
Ofc-Deep & 1.0 & 1.0 & 1.0 & 20 \\\hline
\end{tabular}
}
\end{table}

\section{Discussion and Conclusions}
The average accuracies obtained with JDA and NET using the validation set are comparable to the best accuracies with JDA and NET. 
This empirically validates the model selection proposition. 
However, there is no theoretical guarantee that the parameters selected are the best. 
In the absence of theoretical validation, further empirical analysis is advised when using the proposed technique for model selection.

In this paper, we have proposed the Nonlinear Embedding Transform algorithm and a model selection procedure for unsupervised domain adaptation.  
The NET performs favorably compared to competitive visual domain adaptation methods across multiple datasets. 

This material is based upon work supported by the National Science Foundation (NSF) under Grant No:1116360. 
Any opinions, findings, and conclusions or recommendations expressed in this material are those of the authors and do not necessarily reflect the views of the NSF. 

%

\end{document}